\documentclass[journal]{IEEEtran}

\usepackage{graphicx}
\usepackage{amsmath,amssymb} 
\usepackage{color}
\usepackage{epstopdf}
\usepackage{subfigure}
\usepackage{placeins}
\usepackage{url}
\usepackage{pbox}

\usepackage[pagebackref=true,breaklinks=true,letterpaper=true,colorlinks=true,bookmarks=false]{hyperref}

\usepackage{algorithmic}

\DeclareMathAlphabet{\mathpzc}{OT1}{pzc}{m}{it} 

\newcommand{\etal}{et al.\!}
\newcommand{\eg}{e.g.\!}
\newcommand{\ie}{i.e.\!}

\ifCLASSINFOpdf

\else
\fi

\begin{document}
\title{Boosting Optical Character Recognition: \\A Super-Resolution Approach}

\author{Chao~Dong, Ximei~Zhu,
        Yubin~Deng,
        Chen~Change~Loy,~\IEEEmembership{Member,~IEEE,}
        and~Yu~Qiao~\IEEEmembership{Member,~IEEE}
\thanks{}%
\thanks{C. Dong, Y. Deng and C. C. Loy are with the Department
of Information Engineering, The Chinese University of Hong Kong, Hong Kong.
E-mail: \{dc012, danny.s.deng, ccloy\}@ie.cuhk.edu.hk. 
X. Zhu and Y. Qiao are with the Shenzhen Institutes of Advanced Technology.
E-mail: \{xm.zhu, yu.qiao\}@siat.ac.cn.}
\thanks{}
}

\markboth{ICDAR 2015 Competition on Text Image Super-Resolution}%
{Shell \MakeLowercase{\textit{et al.}}: Bare Demo of IEEEtran.cls for Journals}

\maketitle

\begin{abstract}
Text image super-resolution is a challenging yet open research problem in the computer vision community. In particular, low-resolution images hamper the performance of typical optical character recognition (OCR) systems. In this article, we summarize our entry to the \textit{ICDAR2015 Competition on Text Image Super-Resolution}.
Experiments are based on the provided \textit{ICDAR2015 TextSR dataset}~\cite{Clement2015} and the released \textit{Tesseract-OCR 3.02} system~\cite{ocr}. We report that our winning entry of text image super-resolution framework has largely improved the OCR performance with low-resolution images used as input, reaching an OCR accuracy score of ${77.19\%}$, which is comparable with that of using the original high-resolution images ($78.80\%$).
\end{abstract}

\begin{IEEEkeywords}
super resolution; optical character recognition.
\end{IEEEkeywords}

%
\IEEEpeerreviewmaketitle

\section{Introduction}
%
%
%
%
\IEEEPARstart{O}{ptical} Character Recognition systems aim at converting textual images to machine-encoded text. By mimicking the human reading process, OCR systems enable a machine to understand the text information in images and recognize specific alphanumeric words, text phrases or sentences. Yet, due to the complexity of input contents and variations caused by various sources, accurately recognizing optical characters can be challenging for standard OCR systems. In particular, OCR systems trained on high-resolution (HR) text images may fail on predicting the correctly text in elusive low-resolution (LR) text images. Specifically, LR text images lack of fine details, making it harder for the OCR systems to correctly retrieve the textual information from some commonly acquired OCR features. As such, performing super-resolution on input images is an intuitive pre-processing step towards the goal of accurate optical character recognition.

To perform image super-resolution, unseen pixels need to be predicted under a well-constrained solution space with strong prior. By recovering the HR image from its LR correspondence, pixel values are non-linearly mapped during the process. Following the work~\cite{Dong2014,Dong2014a}, we adopt a Super-Resolution Convolutional Neural Network (SRCNN) approach and train several CNNs targeted on text images. The text image super-resolution CNN framework demonstrates its effectiveness in recovering HR text images. Our main contributions are twofold:

1) We extend the SRCNN approach to the specific text-image domain, and conduct a comprehensive investigation on different network settings (\eg~filter size and number of layers).

2) We conduct model combination to further improve the performance. In the \textit{ICDAR 2015 Competition on Text Image Super-Resolution}\footnote{Competition website: \url{http://liris.cnrs.fr/icdar-sr2015/}}, we achieve the best results~\cite{Clement2015}, which improve the OCR performance by $16.55\%$ in accuracy compared with bicubic interpolation.

\section{Methodology}

\subsection{Super-Resolution Convolutional Neural Network}

The Super-Resolution Convolutional Neural Network (SRCNN)~\cite{Dong2014,Dong2014a} is primarily designed for general single-image super-resolution, and can be easily extended to specific super-resolution topics (\eg~face hallucination and text image super-resolution) or other low-level vision problems (\eg~denoising). The basic framework of SRCNN consists of three convolutional layers without pooling or fully-connected layers. This allows it to accept an LR image $\mathbf{Y}$ of any size (after interpolation and padding) and directly output the HR image $F(\mathbf{Y})$. Each of the three layers is responsible for a specific task. The first layer performs feature extraction on the input image and represents each patch as a high dimensional feature vector. The second layer then non-linearly maps these feature vectors to another set of feature vectors which are conceptually the representation of the HR image patches. The last layer recombines these representations and reconstructs the final HR image. The process can be expressed as the following equations:

\begin{align}
\label{eqn:SRCNN}
F_{i}(\mathbf{Y})&=\max\left(0, W_{i}*\mathbf{Y}+B_{i}\right), i\in\{1,2\}; \\ F(\mathbf{Y})&=W_3*F_{2}(\mathbf{Y})+B_3.
\end{align}
%
where $W_{i}$ and $B_{i}$ represent the filters and biases of the $i$th layer respectively, $F_{i}$ is the output feature maps and '$*$' denotes the convolution operation. The $W_{i}$ contains $n_i$ filters of support $n_{i-1}\times f_i \times f_i$, where $f_i$ is the spatial support of a filter, $n_i$ is the number of filters, and $n_0$ is the number of channels in the input image. Please refer to~\cite{Dong2014a} for more details.

This framework is flexible at the choice of parameters. We can adopt four or more layers with a larger filter size to further improve the performance. For example, in~\cite{Dong2015}, a feature enhancement layer is added after the feature extraction layer to ``denoise'' the extracted features. In the text image domain, we also attempt to use deeper networks to further improve performance.

\subsection{Model combination}

Model combination has been widely used in high-level vision problems (\eg~object detection~\cite{Ouyang2014}). In SRCNN, each output pixel can be regarded as a prediction of a pixel value, thus averaging predictions of different networks could inherently improve the accuracy. Further, as a training based method, the testing results may vary due to different check points and initial values. With model combination, this effect is neutralized, and the results will be more stable and reliable.

We adopt a ``Greedy Search'' strategy to find the best model combination. Suppose we have successfully trained $n$ models. First, we select the model that achieves the highest PSNR (or OCR score) on the validation set. Then we combine its results with that of the $n$ models successively, and get $n$ 2-model combinations. Here, we simply average their output pixel values. The combination with the highest PSNR (or OCR score) is saved as the best 2-model combination. We keep these two models unchanged and find another model from all $n$ models for a second round combination. Similarly, we can identify the best $m$-model combination using these $n$ models. Note that each model can be selected more than once, and the combination of more models is not necessarily better than that of less ones. At last, we choose the best model combination from the $m$ selected model combinations.

\section{Experimentst}

\begin{figure}
\centering
  \includegraphics[width=\linewidth]{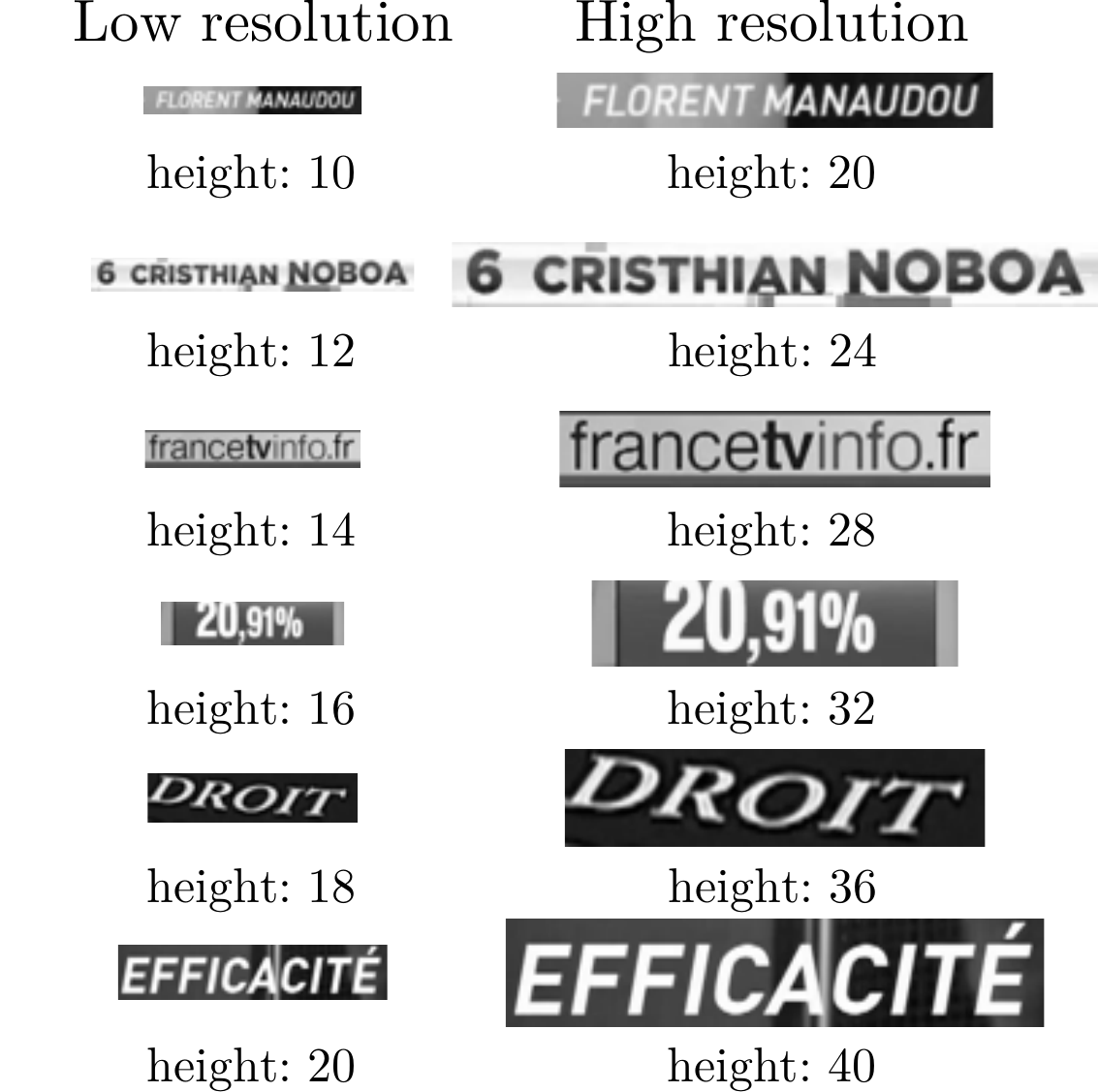}
  \caption{Examples of LR-HR image pairs in the training set.}\label{dataset}
\end{figure}

\textbf{Datasets.} We use the \textit{ICDAR2015 TextSR dataset}~\cite{Clement2015} provided by the \textit{ICDAR 2015 Competition on Text Image Super-Resolution}. The dataset includes a training set and a test set. The training datset consists of 567 HR-LR gray-scale image pairs together with the annotation files. HR and LR images are downsampled by factors of 2 and 4 from the original images that are extracted form French TV video flux. The height of LR images ranges from 9 to 29, and the annotation is realized using standard letters, numbers and 14 special characters. Figure~\ref{dataset} shows several examples of HR-LR image pairs in the training set. The test set contains only 141 LR images, without annotation and HR images. For more details of the dataset, please refer to~\cite{Clement2015}. To evaluate the performance, we select 30 image pairs from the training set as our validation set, thus the training set used in our experiments contains the rest 537 image pairs.

\textbf{Implementation Details.} First, we upsample all LR images by a factor of 2 using Bicubic interpolation. Then the HR and LR images are of the same size in the training set. In the training phase, the ground truth images and LR input samples are prepared as $18\times 18$\footnote{The number 18 is the minimum height of HR images.}-pixel sub-images cropped from the training image pairs. The sub-images are extracted from the original images with a stride of 2 in vertical and 5 in horizontal. In total, the 537 images provide 156,941 training samples. Different from~\cite{Dong2014,Dong2014a}, we cannot remove all borders during training, because the text images are much smaller than generic images. As a compromise, we fix the output size to be $14\time 14$ and pad the input LR images according to the network scales. For example, if there are three layers, then we need to pad $(f_1+f_2+f_3-3)-4$ pixels with zeros. We use the Mean Squared Error (MSE) as the loss function, which is evaluated by the difference between the central $14\times 14$ crop of the ground truth image and the network output. The filter weights of each layer are initialized by drawing randomly from a Gaussian distribution with zero mean and standard deviation 0.001 (and 0 for biases). The learning rate is $10^{-5}$ for the last layer and $10^{-4}$ for the rest layers. All results are based on the check point of 5,000 iterations (about $7.8\times 10^8$ backpropagations).

\subsection{Super-resolution Results of Single Models}

To find the optimal network settings, we conduct four sets of experiments with different filter sizes, number of filters, number of layers and initial values.

\subsubsection{Filter size}
\begin{figure}
\centering
  \includegraphics[width=\linewidth]{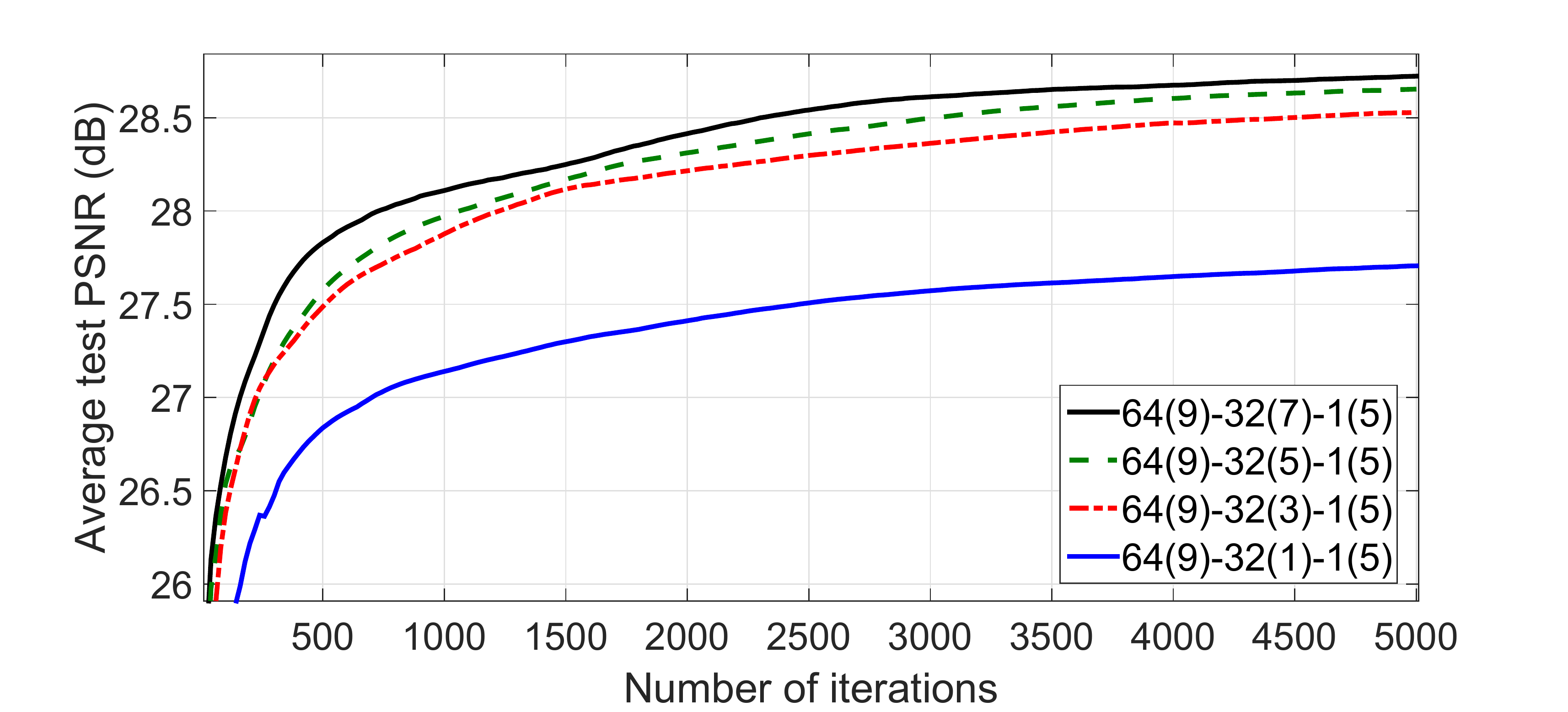}
  \caption{Three-layer networks with different filter sizes.}\label{model1}
\end{figure}

First, we adopt a general three-layer network structure and change the filter size. The basic settings proposed in~\cite{Dong2014} are $f_1=9, f_2=1, f_3=5, n_1=64, n_2=32$ and $n_3=1$, and the network can be denoted as 64(9)-32(1)-1(5). We follow~\cite{Dong2014a} and enlarge the filter size of the second layer $f_2$ to be 3, 5 and 7. Then we have four networks -- 64(9)-32(1)-1(5), 64(9)-32(3)-1(5), 64(9)-32(5)-1(5) and 64(9)-32(7)-1(5). Convergence curves on the validation set are shown in Figure~\ref{model1}. Obviously, using a larger filter size could significantly improve the performance. It is worth noticing that the difference between 64(9)-32(5)-1(5) and 64(9)-32(7)-1(5) is marginal, thus further enlarging the filter size will be of little help.

\subsubsection{Number of Filters}
\begin{figure}
\centering
  \includegraphics[width=\linewidth]{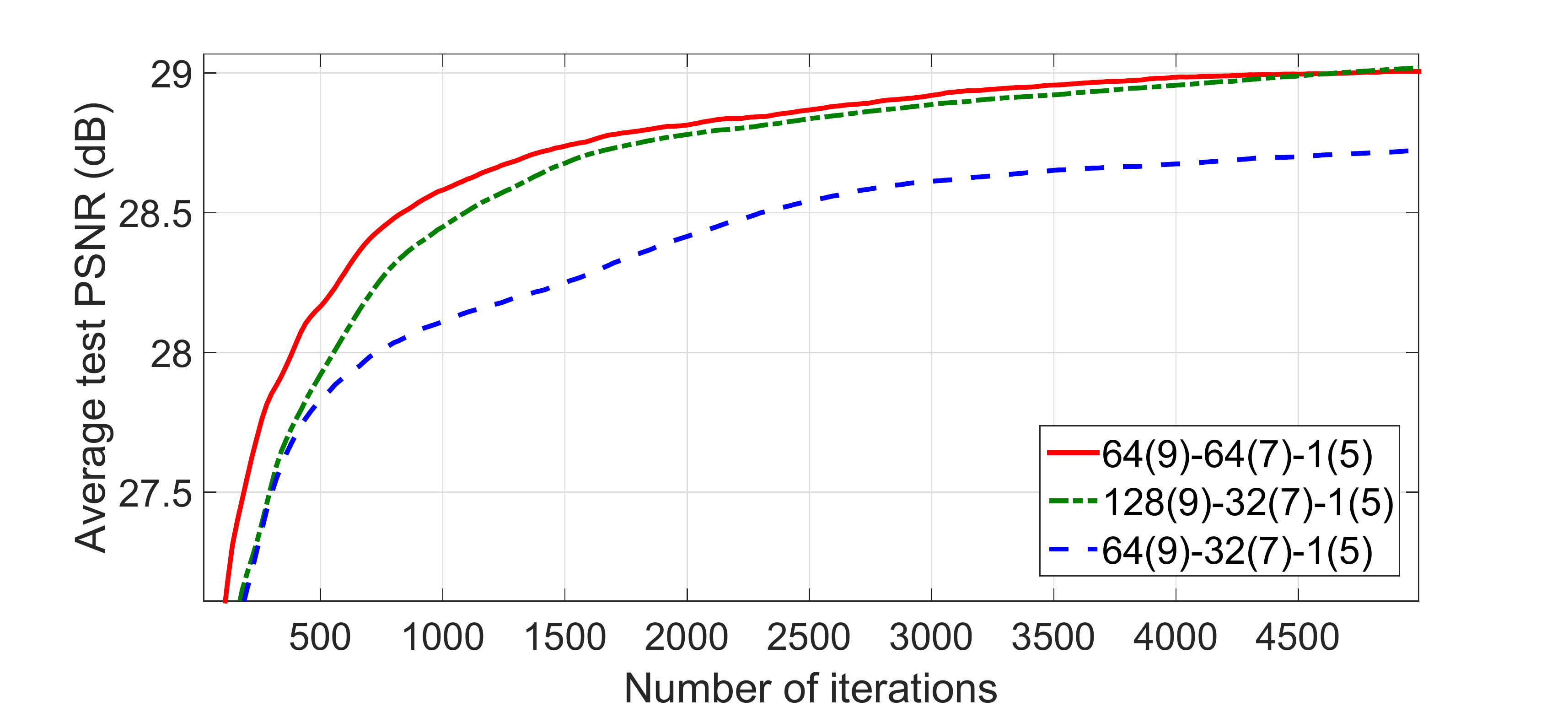}
  \caption{Three-layer networks with different number of filters.}\label{model2}
\end{figure}

We then examine whether the performance benefits from more filters. For doing this, we fix the filter size to be $f_1=9, f_2=7, f_3=5$ and increase the number of filters to be $n_1=128$ and $n_2=64$, separately. The three networks are 64(9)-32(7)-1(5), 128(9)-32(7)-1(5) and 64(9)-64(7)-1(5). From convergence curves shown in Figure~\ref{model2}, we observe that the performance has been pushed much higher. However, when we take a look at the model complexity, 128(9)-32(7)-1(5) and 64(9)-64(7)-1(5) have 211,872 and 207,488 parameters, which are almost double of that for 64(9)-32(7)-1(5) (106,336). Therefore, we wish to investigate other lower-cost alternatives that could still improve the performance.

\subsubsection{Number of Layers}
A reasonable attempt to improve the performance with low cost is to make the network deeper. Here, we adopt a four-layer network structure by embedding another feature enhancement layer as in~\cite{Dong2015}. Further more, we gradually enlarge the filter size to get better results. Overall, we train 8 networks --  64(9)-32(7)-16(1)-1(5), 64(9)-32(7)-16(3)-1(5), 64(9)-32(7)-16(5)-1(5), 64(9)-32(5)-16(5)-1(5), 64(11)-32(9)-16(7)-1(5), 64(11)-32(9)-16(9)-1(5), 64(13)-32(11)-16(9)-1(5) and 64(15)-32(13)-16(11)-1(5).

\begin{figure}
\centering
  \includegraphics[width=\linewidth]{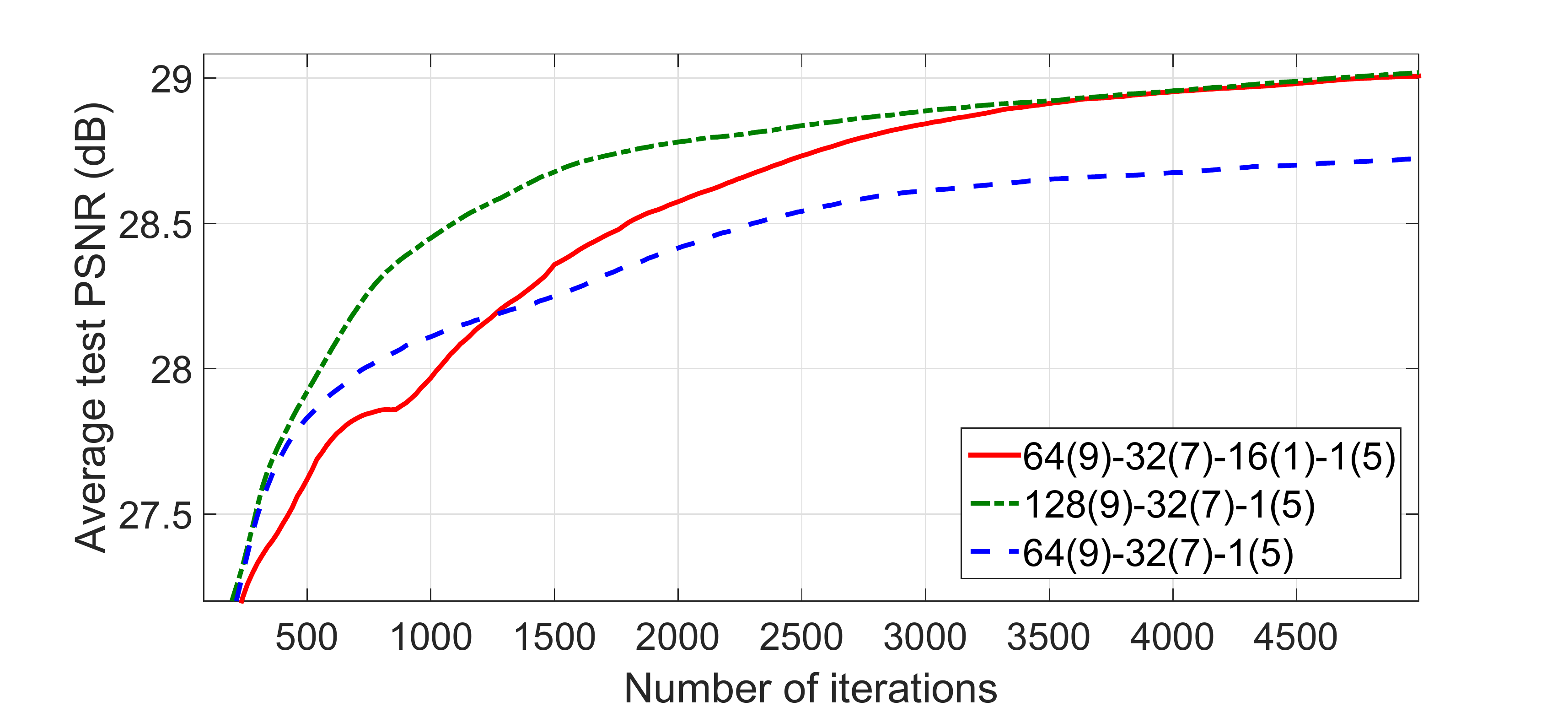}
  \caption{Comparison between three-layer and the four-layer networks.}\label{model3}
\end{figure}

Figure~\ref{model3} reveals the convergence curves of the four-layer network 64(9)-32(7)-16(1)-1(5) and the three-layer networks 128(9)-32(7)-1(5) and 64(9)-32(7)-1(5). Interestingly, the smallest four-layer network 64(9)-32(7)-16(1)-1(5) achieves similar performance as the largest three-layer network 128(9)-32(7)-1(5). Note that the network 64(9)-32(7)-16(1)-1(5) has only 106,448 parameters, which are roughly a half of that for 128(9)-32(7)-1(5) (211,872). This indicates that deeper networks could outperform shallow ones even with less parameters.

\begin{figure}
\centering
  \includegraphics[width=\linewidth]{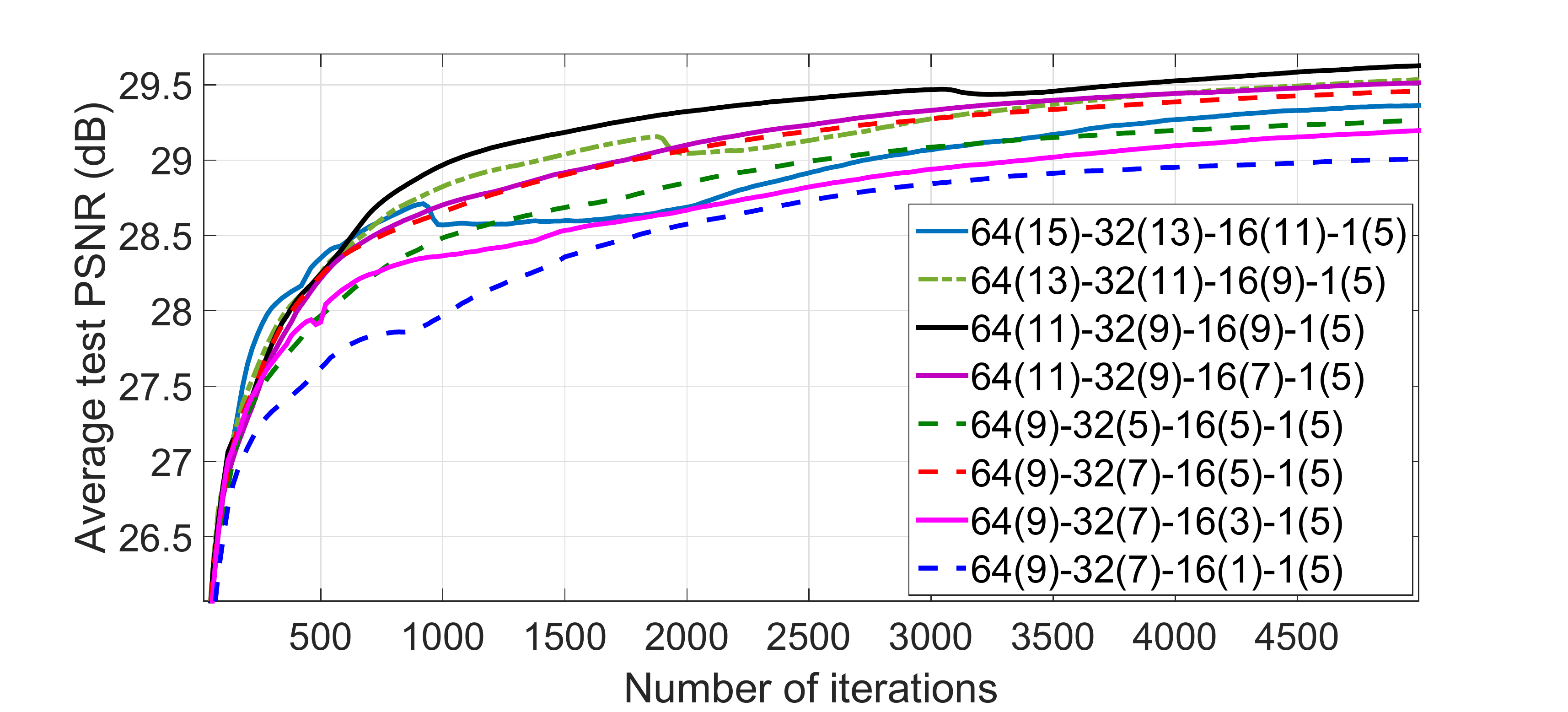}
  \caption{Four-layer networks with different filter sizes.}\label{model4}
\end{figure}

When we enlarge the filter size of the first three layers, the performance improves and reaches a plateau at 64(11)-32(9)-16(9)-1(5)(see Figure~\ref{model4}). Further enlarging the filter size could make the network overfit to the training data (see 64(13)-32(11)-16(9)-1(5) and 64(15)-32(13)-16(11)-1(5)). To avoid overfitting, a larger training set with more diverse images can be introduced, but we stick to the provided training set for competition.

When we try much deeper networks (\ie~five layers), we find that the training is hard to converge even with smaller learning rates. This phenomenon is also observed in~\cite{Dong2014a}, where deeper networks are hard to train on generic images. We may try using other initialization strategies as in~\cite{Dong2015}, but this is out of the scope of this paper.

\subsubsection{Initial values}
\begin{figure}
\centering
  \includegraphics[width=\linewidth]{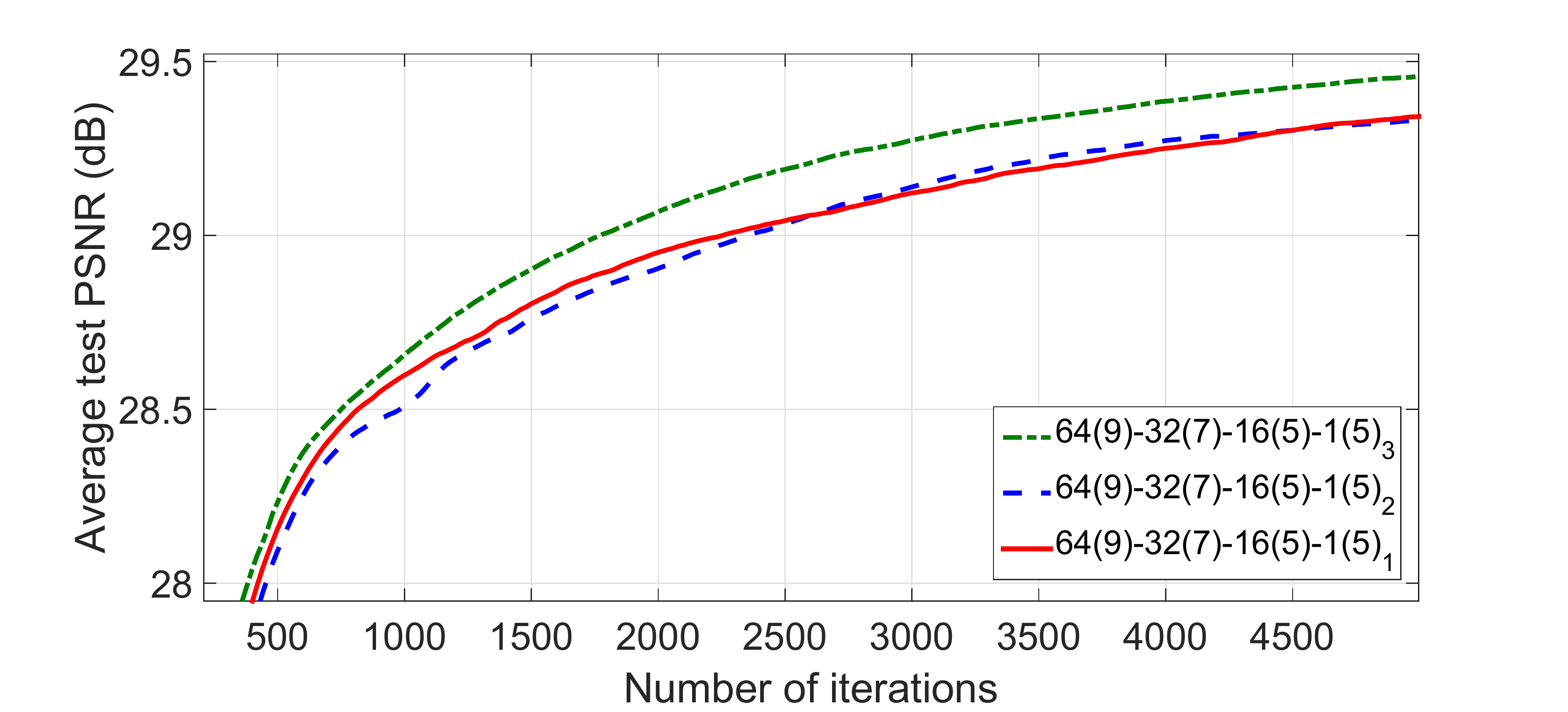}
  \caption{Four-layer networks with different initial values.}\label{model5}
\end{figure}

Then we explore whether the results are affected by different initial values. We train three networks with the same structure 64(9)-32(7)-16(5)-1(5) but different initial values, which are randomly drawn from the same Gaussian distribution for three times. From Figure~\ref{model5}, we could see that the behaviours of three convergence curves are slightly different. This suggests that the results are not unique with different initial values. On the other hand, we could also combine the results of networks with the same structure but different initial values.

\subsection{Model Combination}

To conduct model combination, we select 11 four-layer networks with different filter sizes and initial values. They are 64(9)-32(7)-16(1)-1(5), 64(9)-32(7)-16(3)-1(5), 64(9)-32(7)-16(5)-1(5) (4 networks with different initial values), 64(9)-32(5)-16(5)-1(5), 64(11)-32(9)-16(7)-1(5), 64(11)-32(9)-16(9)-1(5), 64(13)-32(11)-16(9)-1(5) and  64(15)-32(13)-16(11)-1(5). Following the ``Greedy Search'' strategy, we successively obtain 14 best model combinations (from the best single model to the best 14-model combination) evaluated with PSNR. Note that we can also use the OCR score as evaluation criterion, then the results of model combinations could favour a high OCR score.

\begin{figure}
\centering
  \includegraphics[width=0.9\linewidth]{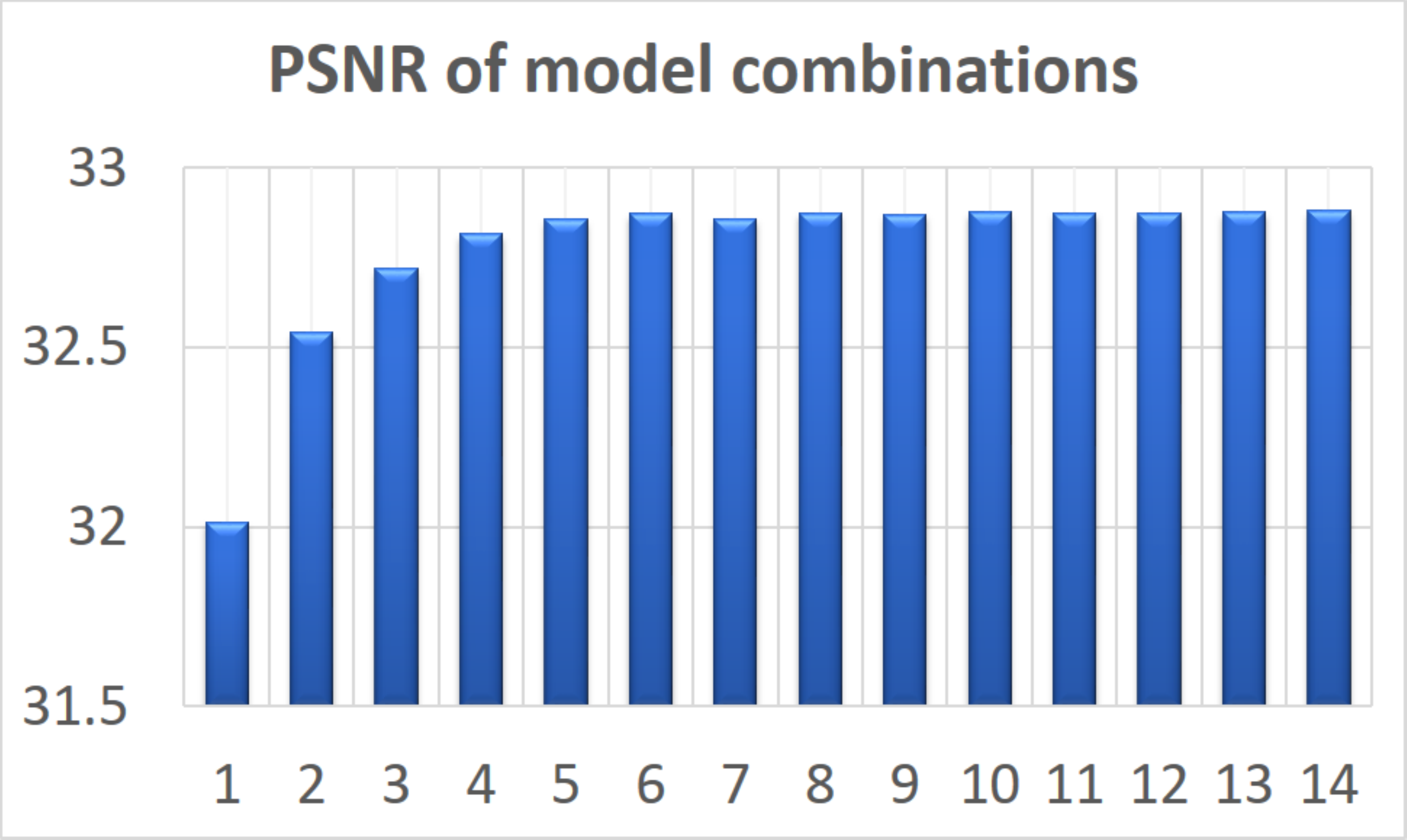}
  \caption{PSNR values of different model combinations.}\label{PSNR}
\end{figure}

The PSNR values\footnote{Note that the PSNR values are not in accord with the convergence curves, since we remove 4-pixel borders during training, but keep them during testing.} of the selected 14 best model combinations are shown in Figure~\ref{PSNR}, from which we have two main observations. First, using model combination could significantly improve the performance, \eg~0.53dB improvement from the best single model to the best 2-model combination. Second, the PSNR values are stable when combining 5 or more models.

\begin{figure}[t]
\centering
\subfigure[]{
  \includegraphics[width=\linewidth]{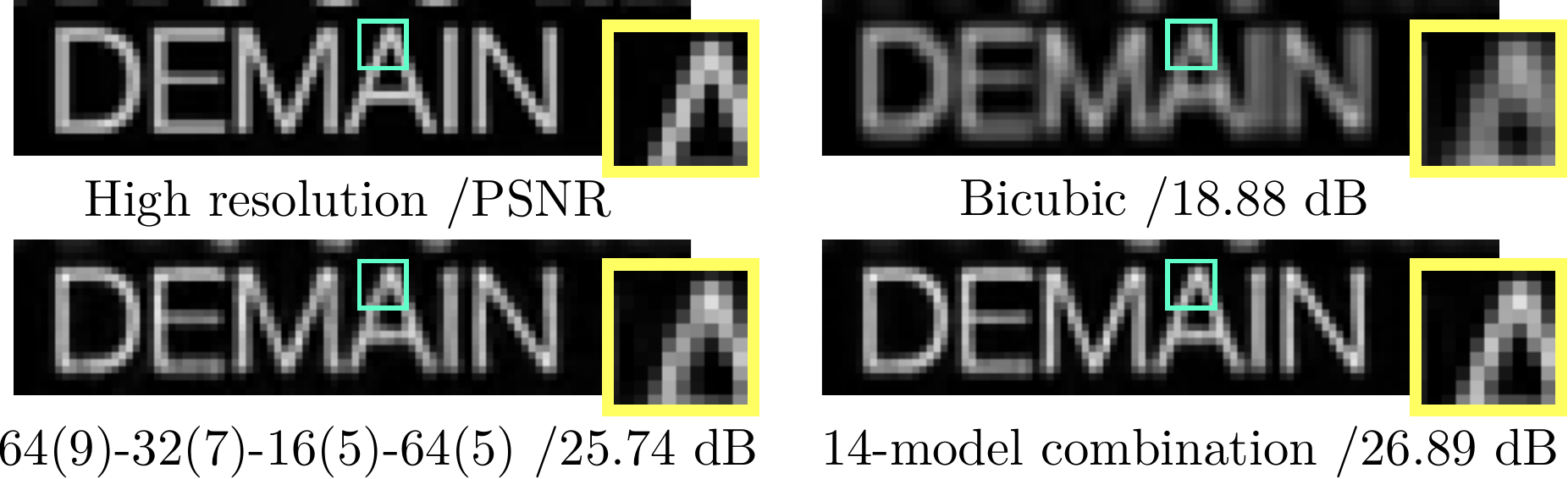}
}
\subfigure[]{
  \includegraphics[width=\linewidth]{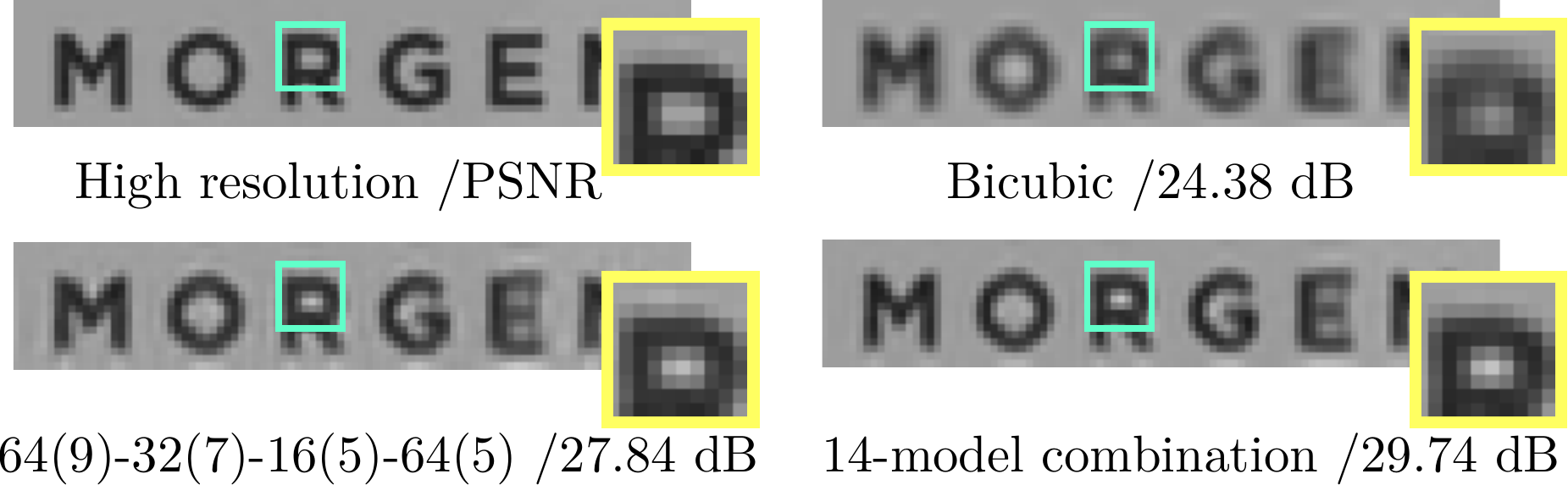}
}
\subfigure[]{
  \includegraphics[width=\linewidth]{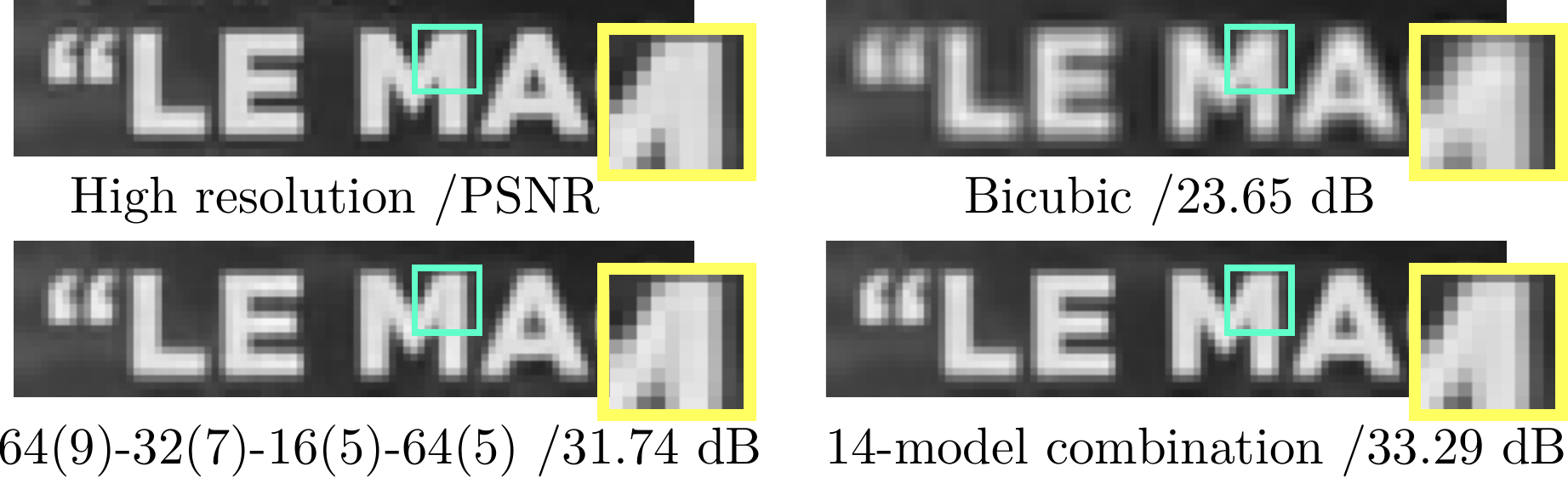}
}
  \caption{Super-resolution results of different methods.}\label{results}
\end{figure}

Figure~\ref{results} shows the super-resolution results of the best single model 64(9)-32(7)-16(5)-1(5) and the best 14-model combination. As can be observed, the super-resolved images are very close to the ground truth HR images.

\subsection{Test Performance}
In the competition, we submitted two sets of results: one favouring the OCR accuracy score (SRCNN-1) and the other favouring the PSNR value (SRCNN-2). The performance is evaluated on the 141 test images with PSNR, RMSE, SSIM and OCR scores by the competition committee. The results are published in~\cite{Clement2015} and listed in Table~\ref{tab:results}. The compared methods are briefly described as follows. The bicubic and lanczos3 interpolation are the baseline methods, the Orange Labs~\cite{Clement2015a} is the internal approach used by the organizer. The Zeyde~\etal~\cite{Zeyde2012} and A+~\cite{Timofte2014} are the representative state-of-the-art super-resolution methods. The Synchromedia Lab~\cite{Moghaddam2010} and ASRS~\cite{Walha2015} are methods of the other two competition teams.
Obviously, our method achieves the best performance both on OCR score (SRCNN-1) and the reconstruction errors (SRCNN-2). The SRCNN-1 improves the OCR performance by $16.55\%$ in accuracy compared with bicubic interpolation. It is worth noticing that the OCR score of using the original HR images is $78.80\%$, which is only $1.61\%$ higher than SRCNN-1. This indicates that our method is extremely effective in improving the OCR accuracy.

\begin{table}
\caption{Results of the \textit{ICDAR2015 Competition on Text Image Super-Resolution}.}\label{tab:results}
\begin{center}
\begin{tabular}{|c|c|c|c|c|}
\hline
Method & RMSE & PSNR & MSSIM & OCR \\
\hline\hline
Bicubic & 19.04 & 23.50 & 0.879 & 60.64 \\
Lanczos3 & 16.97 & 24.65 & 0.902 & 64.36 \\
\hline\hline
Orange Labs~\cite{Clement2015a} & 11.27 & 28.25 & 0.953 & 74.12 \\
\hline\hline
Zeyde~\etal~\cite{Zeyde2012}  & 13.05 & 27.21 & 0.941 & 69.72 \\
A+~\cite{Timofte2014} & 10.03 & 29.50 & 0.966 & 73.10 \\
\hline\hline
Synchromedia Lab~\cite{Moghaddam2010} & 62.67 & 12.66 & 0.623 & 65.93 \\
ASRS~\cite{Walha2015}& 12.86 & 26.98 & 0.950 & 71.25\\
SRCNN-1~\cite{Dong2014,Dong2014a} & 7.52 & 31.75 & 0.980 & \textbf{77.19} \\
SRCNN-2~\cite{Dong2014,Dong2014a} & \textbf{7.24} &\textbf{31.99}& \textbf{0.981} &76.10\\
\hline
\end{tabular}
\end{center}
\end{table}

\section{Conclusions}
In this article, we summarize our attempt of adopting an SRCNN approach in the task of text image super-resolution for facilitating optical character recognition. Our method boosts the baseline OCR performance by a large margin ($16.55\%$). As general image super-resolution has become an increasingly important problem in computer vision, we realize that more comprehensive studies on whether (or to what extend) super resolution further benefits other low-level vision tasks are needed.

\bibliographystyle{splncs03}
\bibliography{long,cnn_sr}

\end{document}